\documentclass{article}

\usepackage{microtype}
\usepackage{graphicx}
\usepackage{subfigure}
\usepackage{booktabs}
\usepackage{hyperref}

\usepackage[accepted]{icml2024}
\usepackage{amsmath}
\usepackage{amssymb}
\usepackage{mathtools}
\usepackage{amsthm}
\usepackage{custom_commands}
\usepackage[capitalize,noabbrev]{cleveref}

\theoremstyle{plain}

\theoremstyle{definition}

\theoremstyle{remark}

\icmltitlerunning{Position: Tensor Networks are a Valuable Asset for Green AI}

\begin{document}

\twocolumn[
\icmltitle{Position: Tensor Networks are a Valuable Asset for Green AI}

\icmlsetsymbol{equal}{*}

\begin{icmlauthorlist}
\icmlauthor{Eva Memmel}{TUD}
\icmlauthor{Clara Menzen}{TUD}
\icmlauthor{Jetze Schuurmans}{comp}
\icmlauthor{Frederiek Wesel}{TUD}
\icmlauthor{Kim Batselier}{TUD}

\end{icmlauthorlist}

\icmlaffiliation{TUD}{Delft Center of Systems and Control, Delft University of Technology, Delft, The Netherlands}
\icmlaffiliation{comp}{Xebia Data, Amsterdam, The Netherlands}

\icmlcorrespondingauthor{Eva Memmel}{e.m.memmel@tudelft.nl}
\icmlcorrespondingauthor{Kim Batselier}{k.batselier@tudelft.nl}

\icmlkeywords{Green AI, Tensor Networks, Tensor Decomposition}

\vskip 0.3in
]
\printAffiliationsAndNotice{}

\begin{abstract}
For the first time, this position paper introduces a fundamental link between tensor networks (TNs) and Green AI, highlighting their synergistic potential to enhance both the inclusivity and sustainability of AI research.
We argue that TNs are valuable for Green AI due to their strong mathematical backbone and inherent logarithmic compression potential.
We undertake a comprehensive review of the ongoing discussions on Green AI, emphasizing the importance of sustainability and inclusivity in AI research to demonstrate the significance of establishing the link between Green AI and TNs.
To support our position, we first provide a comprehensive overview of efficiency metrics proposed in Green AI literature and then evaluate examples of TNs in the fields of kernel machines and deep learning using the proposed efficiency metrics.
This position paper aims to incentivize meaningful, constructive discussions by bridging fundamental principles of Green AI and TNs.
We advocate for researchers to seriously evaluate the integration of TNs into their research projects, and in alignment with the link established in this paper, we support prior calls encouraging researchers to treat Green AI principles as a research priority.
\end{abstract}

\section{Introduction}
More than ever, we have access to data sets throughout almost all science and engineering disciplines. Fueling our economies and shaping our society, data is therefore considered the oil of the 21st century. At the same time, AI algorithms become increasingly powerful to transform large amounts of raw data into valuable information. Consequently, AI development and data availability are deeply intertwined, and they have a common characteristic: both are growing exponentially \cite{wu2021sustainable}. While this wealth of information is opening the doors for extraordinary opportunities, the downside of this development cannot be ignored: AI research on a large scale has adverse side effects on economic, social, and environmental sustainability. Let us consider the example from Strubell et al. that analyzes the energy required for training popular off-the-shelf natural language processing models \yrcite{strubell2019energy}. The training causes CO$_2$ emissions up to \SI{280000}{kg} and cloud computing costs up to 3 million dollars. As a comparison, a person could fly more than 300 times between Amsterdam and New York to emit the same amount of CO$_2$ \cite{myclimate}. 

On another note, Schwartz et al. argued that one of the reasons for the unsustainable development in AI research is unfortunately anchored in what the AI community commonly defines as state-of-the-art results \yrcite{schwartz2020green}, namely focusing on accuracy or similar measures. They claim that to obtain more accurate results, the number of model parameters and hyperparameters is increased, as well as the size of the training data. The result is an exponentially growing demand for compute used to train AI models \cite{schwartz2020green}. This development is not only unsustainable from an environmental and economic point of view but also from a social one: this exponential growth has a large carbon footprint, is expensive, and excludes researchers with fewer resources \cite{schwartz2020green, ahmed2023growing}.

\noindent The problems associated with the unsustainable development of AI have led to a growing awareness in the AI community. Several researchers call for redirecting the focus of AI research by implementing efficiency as an additional benchmark alongside accuracy to assess algorithmic progress \cite{schwartz2020green,strubell2020energy, tamburrini2022ai}. In fact, efficiency has always been the primary criterion to measure algorithmic progress in computer science \cite{knuth1976big,knuth1973fundamental,cormen2022introduction}. Inspired by this, a similar approach can be adapted to AI algorithms. For this purpose, different metrics have been proposed in the literature, e.g.\ estimating the carbon footprint, reporting the energy consumption, or stating the number of floating-point operations (FLOPS) \cite{lacoste2019quantifying,henderson2020towards,lannelongue2021green}. In this context, a new vocabulary has been suggested in \cite{schwartz2020green}. It distinguishes between AI focusing solely on accuracy versus AI considering efficiency and accuracy as equal criteria: Red AI versus Green AI.

With this paper, we want to contribute to the emerging research field into which practices and methods are suitable for Green AI \cite{verdecchia2023systematic, yarally2023uncovering}. 
Green AI methods can roughly be organized into three categories: AI model development, computing infrastructure and system level \cite{wu2021sustainable, kaack2022aligning}. AI model development focuses on reducing energy consumption during data processing, experimentation, training and inference. Computing infrastructure methods aim to, e.g., reduce the environmental impact of computing hardware or data centres. Methods at the system level are, e.g., policy considerations or technology management \cite{wu2021sustainable, kaack2022aligning}. A holistic approach with a portfolio of different methods is critical to achieve a broad effect for Green AI. An in-depth review can be found in several surveys \cite{wu2021sustainable, kaack2022aligning} .

For the first time, we discuss an established tool from multilinear algebra from a Green AI perspective: Tensor Networks (TNs), also called tensor decompositions \cite{kolda2009tensor}. 
TNs fall under the category of AI model development. There, they stand alongside numerous strategies in the category AI model development, such as regularization, automated hyperparameter search \cite{yang2020hyperparameter,bischl2023hyperparameter}, pruning \cite{reed1993pruning, frankle2018lottery}, quantization \cite{gholami2022survey}, physics informed learning \cite{cuomo2022scientific} or drop-out methods \cite{li2022survey}. Most of these methods leverage different properties and do consequently not compete with each other. Instead, they can be combined to provide amplified benefits. To implement a holistic portfolio, we suggest prioritizing understanding the strengths and limitations of individual methods for Green AI.
One strength of TNs is that they approximate data in a compressed format while preserving the essence of the information. In this way, they often achieve logarithmic compression while having the potential to achieve competitive accuracy. This makes TNs precisely the type of tool to apply to large-scale and/or high-dimensional problems: they allow for an efficient and, thus, sustainable way of representing and handling big data \cite{cichocki2014era}. Various promising results regarding accuracy and efficiency, as, e.g.\ \cite{he2018boosted, izmailov2018scalable, richter2021solving, wesel2021large}, show their potential.  

\begin{figure}
    \centering
    \includegraphics[width=0.45\textwidth]{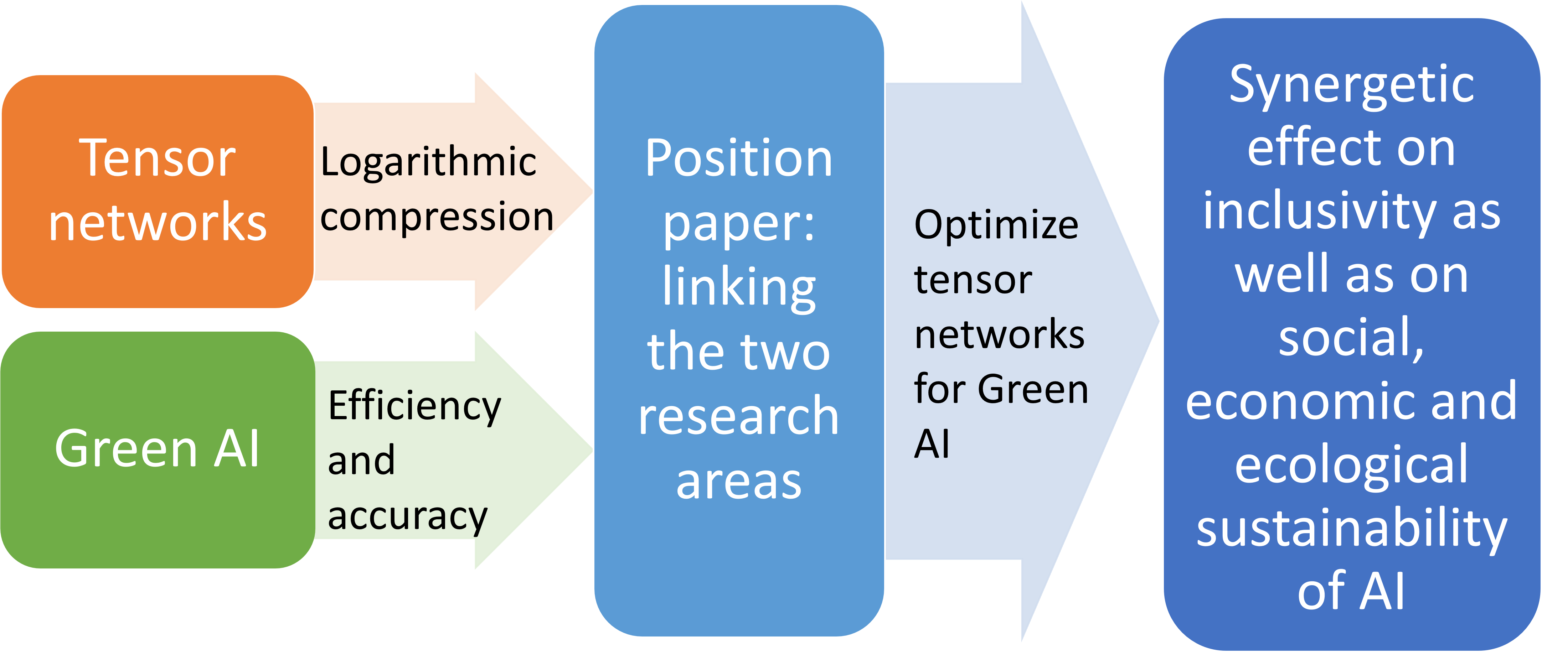}
    \caption{This position paper underscores the importance of connecting the research fields of TNs and Green AI. By establishing this link for the first time, we aim to achieve two primary goals. Firstly, to encourage the TN community to embrace Green AI practices and consciously tailor their TN models for enhanced sustainability. Secondly, to motivate the AI community to adopt Green AI practices and to explore the integration of TNs in their research. We believe that the combined application of Green AI practices and TNs will not only foster the social, economic, and ecological sustainability of AI models but also enhance the inclusivity and diversity of AI research. Ultimately, this synergy is expected to amplify the positive impact of AI research as a whole.}
    \label{fig:overview}
\end{figure}

This potential is well-recognized in the TN community, as evidenced, e.g., by numerous comprehensive surveys \cite{cichocki2016tensor, cichocki2017tensor, panagakis2021tensor, sengupta2022tensor, wang2023tensor}. Although the emphasis and motivation in the TN community so far have not been on promoting the sustainability of AI, the existing technical contributions are a solid foundation for the arguments presented in this position paper. \textbf{This position paper argues that TNs are a valuable asset for Green AI.} To make a substantial impact on Green AI, prioritizing and actively optimizing for sustainability is essential \cite{schwartz2020green}. 
As shown in Figure \ref{fig:overview}, by establishing the link between research on Green AI and TNs for the first time, this position paper aims to encourage interested AI researchers to adopt TNs and Green AI practices in their future work. 
To that end, we provide an economic, social, and environmental analysis of the possibilities and challenges of Green AI as well as the potential of TNs for Green AI. Understanding the impact of TNs on Green AI is not trivial. To the best of the authors' knowledge, this is the first paper that highlights TNs from a Green AI perspective. 
To support our position, the paper is organized as follows: 
In Section \ref{sec:lit}, we underscore the importance of Green AI research by exploring the detrimental effects of the exponentially growing computational demand reported in the literature. We also lay the groundwork for analyzing selected TN examples by reviewing a range of efficiency metrics suggested in Green AI literature. In Section \ref{sec:tensors}, we introduce commonly used TNs and their logarithmic compression potential. In Section \ref{sec:TNAI}, we provide evidence from the literature on how applying TNs in kernel machines and deep learning contexts can lead to efficiency gains. Finally, in Section \ref{sec:conc}, we summarise our findings, demonstrate the benefits of linking TN and Green AI research, critically evaluate our stance and point towards promising future research directions.

\section{Green AI in Related Literature}
 \label{sec:lit}
In their pioneering work, Schwartz et al. point out that AI models are commonly considered state-of-the-art when they achieve greater accuracy (or similar measures) than previously reported \yrcite{schwartz2020green}. 
To incentivize competition and thus accelerate AI innovation, results are made public on leaderboards based on accuracy metrics. Schwartz et al. assert that beating the state-of-the-art algorithm is often achieved by at least one of the following three aspects: more extensive data sets, a more complex model, which is highly correlated with the number of parameters, or more extensive hyperparameter experiments \yrcite{schwartz2020green}. 
This results in an exponentially growing demand for compute to train AI models \cite{OpenAICompute}, requiring an unsustainable amount of hardware, energy, and computational time. 

We review the related literature in two parts. First, we summarise how existing literature views the negative impact of increasing computation on sustainability in general and AI progress in particular. Second, we highlight metrics proposed in the literature for measuring efficiency or estimating the environmental impact of AI models.  

\subsection{The Negative Impact of Growing Compute on Sustainability and AI Progress}
Sustainability is based on three fundamental, intersecting dimensions: economic, social, and environmental \cite{purvis2019three}. 
Research indicates that the rapidly increasing computational demands in AI have the potential to challenge these three dimensions. The implications of this are explored in the subsequent discussion.

As large computations can have a hefty price tag \cite{strubell2019energy,strubell2020energy}, they negatively impact both the economic and social dimensions of sustainability. 
Concerning the \textbf{economic} dimension, linear gains in accuracy contrast with an exponentially growing amount of compute \cite{schwartz2020green}. Thus, diminishing returns contrast with increasing costs associated with e.g.\ cloud computing and hardware (such as CPU, GPU, and TPU).
In the \textbf{social} dimension, high costs can create barriers for researchers with fewer resources to participate in computationally expensive AI development. The barriers for researchers with fewer resources (academia typically has fewer resources than industry) are reflected in the drop of academic contributions to large-scale AI results from 89\% in 2010 to 4\% in 2021, with the rest claimed by industry \cite{ganguli2022predictability, ahmed2023growing}.
Beyond the financial aspect, being dependent on external cloud compute providers can be problematic, too \cite{ahmed2023growing}. For example, computations may need to be performed on-site in applications where privacy or security-relevant data is handled. Therefore, state-of-the-art computations with sensitive data may only be possible if enough financial resources are available to provide for large amounts of expensive hardware.
Concerning the \textbf{environmental} aspect, several studies show that AI research causes a substantial carbon footprint \cite{strubell2020energy, brown_language_2020, dodge2022measuring}.
The emissions are attributable to operational emissions, associated, e.g., with cloud computing, and embodied emissions, associated, e.g., with hardware. 
In case operational emissions are decreased by relying on electricity with a low carbon intensity, embodied emissions make up for the largest share \cite{gupta2021chasing}.
In fact, a recent study shows that more than 50\% of Meta's emissions are attributed to embodied costs \cite{wu2021sustainable}.

Beyond sustainability, an exponentially growing computational need can cause an additional problem. When computational needs cannot be met anymore, they will negatively impact progress and innovation in AI. 
Until now, progress and innovation in AI strongly rely on the increase of available computational resources \cite{thompson2020computational}. 
According to forecasts, the number of transistors on a microchip doubling every two years (Moore's Law), will reach its limits \cite{leiserson2020there}. 
Furthermore, its two-year doubling period has already been overtaken by the exponentially growing need for compute, which is doubling every 3.4 months \cite{OpenAICompute} with a total increase of  $300,000$ times between 2012 and 2018 \cite{OpenAICompute}. 
Consequently, the growing need for compute will ultimately limit AI progress and innovation, especially in computationally expensive fields \cite{strubell2019energy,thompson2020computational}. 
Another aspect is that exponentially growing demand for compute can compromise reproducibility. The AI community already has a growing awareness of its importance; see, e.g.,\ the ML Reproducibility Challenge \cite{paperswithcode}.

To summarize, reducing compute can tackle many of the problems mentioned above. One suggested direction is for AI researchers to redirect their focus toward Green AI and redefine what a state-of-the-art model entails. 
First, it is suggested that accuracy and efficiency be considered equally important when measuring progress in AI. 
Second, an analysis of the trade-off between performance and computational resources used should be included in AI research. 
It can be concluded that AI algorithms with higher efficiency will, therefore, positively impact both sustainability and AI progress \cite{strubell2019energy,schwartz2020green}. 
In addition, reporting on efficiency metrics has other benefits: it will raise awareness and incentivize progress in efficiency \cite{henderson2020towards,tamburrini2022ai}. 

\subsection{Efficiency Metrics Suggested in Literature}
\label{sec:effmetric}
So far, there is a tendency at major AI venues to report accuracy rather than efficiency or both metrics \cite{schwartz2020green}. Reporting efficiency for AI models can present challenges, and currently, there is no standardized approach for this task in the AI community \cite{hernandez2020measuring}. A thorough understanding of their benefits and limitations is essential to employ efficiency metrics effectively. Consequently, we will discuss various evaluation criteria for efficiency that are considered in the literature, namely CO$_2$ equivalent (CO$_2$e) emissions, electricity usage,  floating point operations (FLOPS), the big $\mathcal{O}$ notation, elapsed runtime and the number of parameters.

An appealing criterion to measure efficiency is CO$_2$e emissions since they are directly related to climate change. Disadvantages include that they are challenging to measure and are influenced by factors that do not account for algorithmic optimization, e.g.\ hardware, the carbon intensity of the used electricity, and time as well as the location of the compute~\cite{schwartz2020green,strubell2020energy}. 
An alternative criterion independent of time and location is to state the electricity usage in kWh. It can be quantified because GPUs commonly report the amount of consumed electricity. 
However, it is still hardware-dependent.
There are several websites and packages available to compute the CO$_2$e and electricity usage of algorithms \cite{lacoste2019quantifying,lottick2019nergy,anthony2020carbontracker,henderson2020towards,lannelongue2021green,codecarbon}.

A hardware-independent criterion is the number of FLOPS required to train the model. On the one hand, they are computed analytically and facilitate a fair comparison between algorithms \cite{hernandez2020measuring,schwartz2020green}. On the other hand, they do not account, e.g.,\ for optimized memory access or memory used by the model \cite{henderson2020towards,schwartz2020green}. Next to giving the absolute number of FLOPS, it is possible to additionally report e.g.\ a FLOPS-based learning curve~\cite{hernandez2020measuring}. Several packages are available to compute FLOPS, such as flops-counter, torchstat, pytorch-0pCounter or GFlops \cite{torchstat, flops-counter.pytorch, GFlops, pytorch-OpCounter}.\\
A commonly used criterion in computer science is the \mbox{big $\mathcal{O}$} notation \cite{knuth1976big}. It is used to report an algorithm's storage complexity and computational cost. 
While the \mbox{big $\mathcal{O}$} notation may be impractical for AI practitioners due to the strong influence of application-specific termination criteria on run time, it offers significant theoretical benefits. It allows for easy comparison of algorithm sections and required storage complexities, making it a valuable tool in theoretical analysis. \\
The elapsed runtime is easy to measure but highly dependent on the used hardware and other jobs running in parallel \cite{schwartz2020green}. The number of parameters, whether they are learnable or total, is a widely used measure of efficiency. It is reassuringly agnostic to hardware and takes into account the memory needed by the model. However, it is essential to note that different model architectures can result in varying workloads for the same number of parameters.

\section{Essentials of Tensor Networks for Green AI}
\label{sec:tensors}
As mentioned earlier, increasing the model parameters, hyperparameters, and training data size can boost an AI model's performance. 
TNs can handle the resulting large-scale or high-dimensional data objects.
This paper builds upon the evidence for the compression potential of TNs provided in prior work. By establishing the link between TNs and Green AI explicitly for the first time, we aim to inspire future research that utilizes TNs beyond efficiency improvement but optimizes TNs as a part of a holistic portfolio for Green AI.\\ 
Alongside several reviews on TNs \cite{kolda2009tensor,cichocki2016tensor}, several surveys and technical contributions discuss the broad applicability and successful implementation of TNs in AI. The work referenced below is a small selection of existing work for TNs in AI; a complete overview is well beyond the scope of this paper. 
The surveys \cite{cichocki2016tensor, cichocki2017tensor, Yuwang2019survey} give a broad overview of the applications of TNs, covering data preprocessing, supervised and unsupervised learning, regression and classification tasks, Gaussian Processes (GPs), kernel machines or deep learning, among others. A detailed overview of the various applications of TNs in neural networks (NNs) is provided by the survey \cite{wang2023tensor}, covering, for example, Convolutional Neural Networks (CNNs), Recurrent Neural Networks (RNNs) and transformers. 
Collaborative filtering, mixture and topic modelling with TNs are discussed in~\cite{Sidiropoulos2017tensor}. Finally, in \cite{SIGNORETTO2011861}, a framework for kernel machines with TNs is discussed. \\
While the referenced sources are excellent for detailed technical knowledge, our paper primarily focuses on the potential of TNs for Green AI. Before applying TNs to AI is discussed in Section \ref{sec:TNAI}, this section explains the foundational concepts of TNs.
\subsection{From Vectors and Matrices to Tensors and then Tensor Networks}
\label{sec:tensorization}
\begin{figure}
    \centering
    \includegraphics[width=0.5\textwidth]{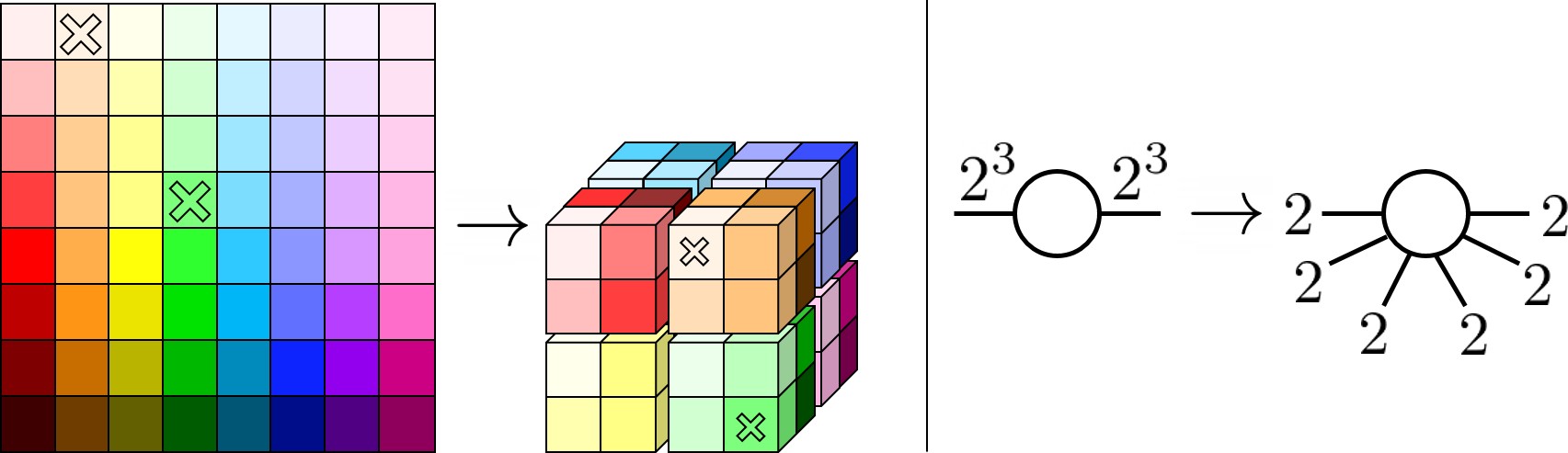}
    \caption{Left: tensorization of a matrix of size $2^3\times2^3$ into a sixth-order tensor of size $2\times2\times2\times2\times2\times2$. The two crosses illustrate the organization of the entries. Based on Fig. 2 of \cite{cichocki2015tensor}. Right: tensorization in diagram notation. The 2 edges of the node representing the 8-by-8 matrix become 6 edges sticking out the node representing the sixth-order tensor.}
    \label{fig:largevec}
\end{figure}

Multidimensional arrays, widely known as tensors, are a generalization of vectors and matrices to higher orders\footnote{In the tensor community, order commonly refers to the number of indices: For example, a third-order tensor has three indices and can be visualized in a cube \cite{kolda2009tensor}.}. The primary data structures in various fields are large vectors or matrices rather than tensors, which could imply that TNs might not be suitable for many standard applications.
However, there is a solution to the problem described above: vectors and matrices can be rearranged into tensors by a procedure called tensorization \cite{Oseledets2010,khoromskij2011dlog}. The row and column size are factorized into multiple factors and then reshaped into a tensor.
A small tensorization example is illustrated in Figure \ref{fig:largevec}, where a matrix of size $2^3\times2^3$ is transformed into a sixth-order tensor, visually represented as a cube of smaller cubes. 
Tensorization enables the application of TNs to vectors and matrices through systematic index bookkeeping. The two 'X' shapes on the left side of Figure \ref{fig:largevec} illustrate this method of organizing entries. The essential requirement to preserve information during tensorization is meticulous indices tracking. For instance, the element labelled with the light orange 'X' in Figure \ref{fig:largevec}, which is represented as $(1,\,2)$ in matrix format, transforms into $(1,\,1,\,1,\,1,\,2,\,1)$ in the tensorized form.
 Due to this bookkeeping approach, the impact of the newly imposed structure is expected to be minimal. In practice, the tensorization is simply performed as a \texttt{reshape()} operation, which does not introduce relevant computational costs.
To simplify the depiction of higher-order objects, it is possible to use the diagram notation shown on the right side of Figure \ref{fig:largevec}. 
In this notation, a matrix or a tensor is depicted as nodes with as many edges sticking out as its number of orders. 
In the following subsection, we introduce commonly used TNs. 

\subsection{Commonly Used Tensor Networks} \label{sec:deftn}
A tensor can be expressed as a function of simpler tensors that form a TN, also called tensor decomposition. The idea of a TN originates in the generalization of the Singular Value Decomposition (SVD) to higher orders. In many applications, TNs can represent data in a compressed format with a marginal loss of information because of correlations present in the data \cite{cichocki2016tensor,cichocki2017tensor}.  

\begin{figure}
    \centering
    \includegraphics[width=0.45\textwidth]{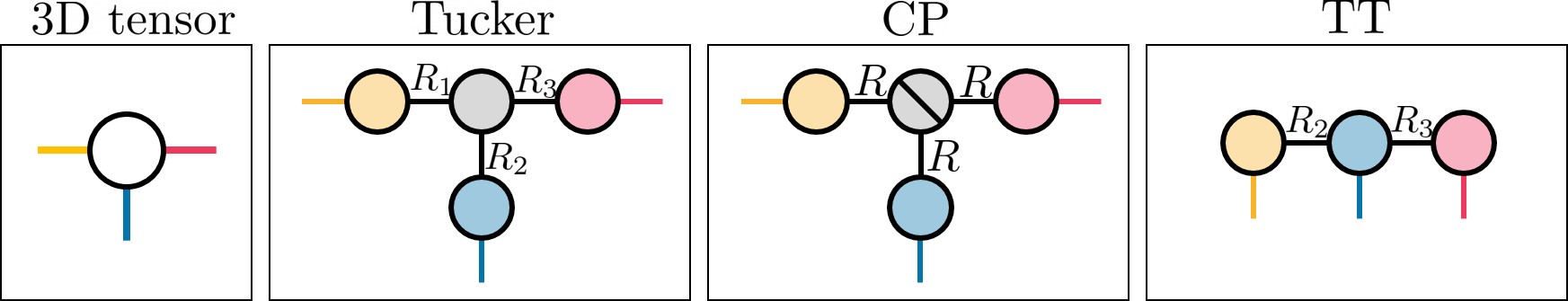}
    \caption{Graphical depiction of commonly used TNs for a third-order tensor. Connected edges are indices that are being summed over. The CP decomposition is a special case of Tucker, where the core tensor is diagonal. This is shown by the diagonal in the node.}
    \label{fig:TDs}
\end{figure}

In literature, the most commonly used TNs include the Canonical Polyadic (CP) \cite{Carroll1970,Harshman1970}, the Tucker \cite{Tucker1966}, and the Tensor Train (TT) decomposition \cite{oseledets_tensor-train_2011}. Without loss of generality, in this subsection, we will treat a third-order tensor $\Yt$ to introduce the TNs mentioned above.
The \textbf{CP decomposition} \cite{Carroll1970,Harshman1970} of $\Yt\in\mathbb{R}^{I_1\times I_2\times I_3}$ consists of a set of factor matrices $\A,\B,\C \in \mathbb{R}^{I_j \times R}$, $j=1,2,3$ and a weight vector $\boldsymbol\lambda\in\mathbb{R}^{R\times 1}$. Elementwise, $\Yt$ can be computed from
\begin{equation}
    y_{i_1i_2i_3} = \sum_{r=1}^R \lambda_r \; a_{i_1r} \; b_{i_2r} \; c_{i_3r}.
\end{equation}
The scalars $a_{i_1r}, b_{i_2r}, c_{i_3r}$ are the entries of the three factor matrices $\A,\B,\C$, $\lambda_r$ is the $r$-th entry of $\boldsymbol\lambda$, $R$ denotes the rank of the decomposition.
The \textbf{Tucker decomposition} \cite{Tucker1966} of $\Yt\in\mathbb{R}^{I_1\times I_2\times I_3}$ consists of a 3-way tensor $\mathbfcal{G}\in\mathbb{R}^{R_1\times R_2 \times R_3}$, called the core tensor, and a set of matrices $\A,\B,\C \in \mathbb{R}^{I_j \times R_j}$, $j=1,2,3$. Elementwise, $\Yt$ can be computed from
\begin{equation}
y_{i_1i_2i_3} =\sum_{r_1=1}^{R_1} \sum_{r_2=1}^{R_2} \sum_{r_3=1}^{R_3} g_{r_1r_2r_3} \; a_{i_1r} \; b_{i_2r} \; c_{i_3r}.
\end{equation}
The scalars $a_{i_1r}, b_{i_2r}, c_{i_3r}$ are the entries of the three factor matrices $\A,\B,\C$, $g_{r_1r_2r_3}$ is the $(r_1r_2r_3)$-th entry of $\Gt$ and $R_1, R_2, R_3$ denote the ranks of the decomposition. 
The \textbf{TT decomposition} ~\cite{oseledets_tensor-train_2011} of $\Yt\in\mathbb{R}^{I_1\times I_2\times I_3}$ consists of a set of three-way tensors $\Yt^{(j)}\in \mathbb{R}^{R_j\times I_j \times R_{j+1}}$, $j=1,2,3$ called TT-cores. Elementwise, $\Yt$ can be computed from
\begin{equation} \label{eq:tt}
y_{i_1i_2i_3} = 
\sum_{r_1=1}^{R_1} \sum_{r_2=1}^{R_2} \sum_{r_3=1}^{R_3} \sum_{r_4=1}^{R_4} y^{(1)}_{r_1i_1r_2}\;y^{(2)}_{r_2i_2r_3}\;y^{(3)}_{r_3i_3r_4},
\end{equation}
where $R_1,R_2,R_3,R_4$ denote the ranks of the TT-cores and by definition $R_1= R_4=1$.
Figure \ref{fig:TDs} shows a diagram depicting the CP, Tucker, and TT decomposition for the case of a third-order tensor. Connected edges are indices being summed over, whereas the number of free edges corresponds to the order of the tensor. 

When a TN approximates a tensor,  the chosen ranks are pivotal. These ranks act as hyperparameters and are crucial for determining the approximation's accuracy, necessitating careful tuning. The usual goal is to balance compression and accuracy effectively. It is essential to select ranks that are low enough to ensure effective compression but still high enough to maintain the desired level of accuracy of the approximation.

\subsection{Logarithmic Compression Potential of Low-Rank Tensor Networks as an Enabler for Green AI}
The low-rank approximation of TNs is powerful for two key reasons.
First, computations can be performed in the compressed low-rank format, typically executed at the level of individual TN components such as CP factor matrices or TT cores. Due to these mathematical properties, TNs can efficiently compress both data and model parameters \cite{cichocki2016tensor, cichocki2017tensor}.
Second, low-rank TNs can transform an exponential complexity into a linear complexity with minimal loss of information \cite{cichocki2016tensor, cichocki2017tensor}. We call this type of transformation logarithmic compression. 
The inherent logarithmic compression capability of TNs is a key reason we present them as a powerful tool for Green AI in this position paper.

Considering a CP decomposition for a given tensor with $I^D$ elements, the number of elements in its rank-$R$ decomposition are $\mathcal{O}(RID)$, thus linear in $D$. Assuming a uniform rank $R$ in a TT decomposition yields a storage complexity of $\mathcal{O}(R^2ID)$ elements. Consequently, both decompositions achieve logarithmic compression. 
The Tucker decomposition, on the other hand, still scales exponentially with $D$: a given tensor with $I^D$ elements requires $\mathcal{O}(R^D)$ in the Tucker format. Tucker decompositions can still achieve significant compression if $R\ll I$. Figure \ref{fig:compressionPlot} illustrates the effects of different decompositions for a numerical example.

\begin{figure}
\centering
\resizebox{0.4\textwidth}{!}{%
\begin{tikzpicture}
\begin{axis}[ymode=log, major grid style={black!50}, xlabel={number of modes $D$}, ylabel={storage complexity}, xtick={1,2,3,4,5,6,7,8}, xticklabels={2,4,6,8,10,12,14,16},legend columns=-1, legend style={at={(0.45,1.15)},anchor=north},style = {very thick}, grid=both]
\addlegendimage{empty legend}
    \addplot[color={purple}, mark={*}]
        coordinates {
            (1,100^2) 
            (2,100^4)
            (3,100^6) 
            (4,100^8) 
            (5,100^10) 
            (6,100^12)
            (7,100^14)
            (8,100^16) 
        }
        ;
    \addplot[color={yellow}, mark={*}]
        coordinates {
            (1,10^2*100) 
            (2,10^4*100)
            (3,10^6*100) 
            (4,10^8*100) 
            (5,10^10*100) 
            (6,10^12*100)
            (7,10^14*100)
            (8,10^16*100) 
        }
        ;
    \addplot[color={cyan}, mark={*}]
        coordinates {
            (1,1*100*10^2*2) 
            (2,2*100*10^2*2) 
            (3,3*100*10^2*2)
            (4,4*100*10^2*2) 
            (5,5*100*10^2*2) 
            (6,6*100*10^2*2)
            (7,7*100*10^2*2)
            (8,8*100*10^2*2) 
        }
        ;
    \addplot[color={green}, mark={*}]
        coordinates {
            (1,1*100*10*2) 
            (2,2*100*10*2) 
            (3,3*100*10*2)
            (4,4*100*10*2) 
            (5,5*100*10*2) 
            (6,6*100*10*2)
            (7,7*100*10*2)
            (8,8*100*10*2) 
        }
        ;
 \addlegendentry{\vspace{-.6cm} Decomposition:}
   \addlegendentry{No}
   \addlegendentry{Tucker}
      \addlegendentry{TT}
   \addlegendentry{CP}
\end{axis}
\end{tikzpicture}
}
\caption{Demonstrating the impact of Tucker, TT, and CP decompositions on a $I^D$ tensor with $I=100$ and $R=10$. Without decomposition, the tensor's storage complexity increases exponentially with $D$. The Tucker decomposition yields a slower exponential growth, while CP and TT decompositions grow linear in $D$.}\label{fig:compressionPlot}
\end{figure}
\section{Applications of Tensor Networks to AI}
\label{sec:TNAI}
This position paper promotes the strategic application of TNs to develop more sustainable AI algorithms. 
So far, we have established that TNs have broad applicability for various AI model architectures, learning paradigms, and tasks. When optimally leveraged, they have the potential to meet the growing computational needs in AI due to their logarithmic compression capability while maintaining accuracy. 
Thanks to efficient bookkeeping, TNs are versatile. They are adept at processing not only tensor-formatted data but also vectors and matrices, making them suitable for a wide range of AI applications.

This section aims to enrich existing technical surveys on TNs by injecting a distinct Green AI perspective supported by carefully curated examples from kernel machines and deep learning. We delve into a typical supervised learning scenario to illustrate the practical integration of TNs into AI algorithms. 
Given a set of independent and identically distributed input/output pairs $\x_n$ and $\y_n$, a supervised learning problem can be described as minimizing the measure of loss $L$ and a regularization term~$R$ 
\begin{align}
    \min_\w  \; \frac{1}{N} \sum_{n=1}^{N} L(\y_n,f(\x_n\mid\w)) + R(\w),
    \label{eq:regression}
\end{align}
where $f(\x\mid\w)$ is a nonlinear function parameterized by the weights $\w$.
To incorporate TNs, $f(\x\mid\w)$ is parametrized with low-rank TNs.
We will explore two specific implementations of this parametrization: a kernel machine and a NN. 
In discussing these models, we will highlight efficiency improvements and the associated changes in accuracy as documented in the literature, demonstrating how TNs enhance the efficiency of AI algorithms. 
The insights gained from these examples showcase the practical efficacy of TNs in advancing sustainable AI and underscore their potential for broader application across various AI domains.

\subsection{Tensor Networks in Kernel Machines}
Kernel machines, such as Gaussian processes \cite{Rasmussen2006} and support vector machines \cite{cortes1995support}, can be universal function approximators \cite{hammer2003note}.
While they have shown equivalent or superior performance compared to NNs \cite{lee2017deep,garriga2018deep,novak2018bayesian}, they scale poorly for high-dimensional or large-scale problems. 
A single-output kernel machine is given by
\begin{equation}
    \label{eq:model1}
    f(\x\mid\w) = \langle \boldsymbol\varphi(\x),\w \rangle, 
\end{equation}
where $\boldsymbol\varphi(\cdot)$ is a feature map, $\w$ is a weight vector and $\langle\cdot,\cdot\rangle$ denotes the inner product. 
A common choice is to represent $\boldsymbol\varphi(\cdot)$ as a Kronecker product of $D$ regressors, computed from a chosen number $I$ basis functions. 
The adoption of a Kronecker product structure can lead to significant storage and computational complexity, resulting from the exponential increase of the number of basis functions $\boldsymbol\varphi(\cdot) \in \mathbb{R}^{I^D}$ and parameters in $\w  \in \mathbb{R}^{I^D}$. 
The issue of this exponential growth can be completely alleviated by imposing a TN structure on both $\boldsymbol\varphi(\x)$ and $\w$, leveraging the logarithmic compression potential of TNs. In other words, the basis functions $\varphi(\x)$ and weight vector $\w$ are never explicitly calculated. Instead, their TN representations are used throughout training and inference. For kernel machines with TNs, the following basis functions have been explored in the literature:
polynomial basis function are treated in \cite{rendle_factorization_2010,blondel_higher-order_2016,batselier_constructive_2017}, pure-power-$1$ polynomials in \cite{novikov_exponential_2017}, lagged timeseries in \cite{batselier2017tensorALS}, pure-power-$I$ polynomials in \cite{chen_parallelized_2018} and B-splines in \cite{karagoz_nonlinear_2020}. 
The use of trigonometrical basis functions is described in \cite{stoudenmire_supervised_2016}, and Fourier features in \cite{wahls_learning_2014,kargas_supervised_2021,wesel2021large}. 
Wesel \& Batselier, for example, achieved superior accuracy results with TNs on a laptop compared to those previously obtained by an alternative method on a cluster \yrcite{wesel2021large}.\\
Employing TNs can significantly reduce a model's memory requirements. For example, applying CP decomposition reduces the storage complexity from $\mathcal{O}(I^D)$ down to $\mathcal{O}(DIR)$, and the total computational runtime for training from $\mathcal{O}(NI^{2D}+I^{3D})$ down to $\mathcal{O}(DN(IR)^2 + D(IR)^3)$ \cite{wesel2021large}. Significant computational savings are attained by performing both training and inference computations at the level of a single factor matrix with dimensions $I \times R$. The key insight is the transformation of the dependency on $D$ from an exponential to a linear scale, a change that substantially boosts efficiency without compromising accuracy.

\subsection{Tensor Networks in Deep Learning}
Deep learning has achieved state-of-the-art performance in many fields, such as 
computer vision~\cite{krizhevsky_imagenet_2012,he_deep_2016} and Natural Language Processing
(NLP) \cite{devlin_bert_2019,brown_language_2020}. The success, however, comes at a cost: models in deep learning are large and require a lot of compute \cite{strubell2019energy,strubell2020energy}. 
NNs have been made more efficient with TNs for a variety of application fields,
including computer vision~\cite{jaderberg_speeding_2014,lebedev_speeding-up_2015,kim_compression_2016} 
and NLP~\cite{ma_tensorized_2019,hrinchuk_tensorized_2020, abronin2024tqcompressor}. \\
We address the learning problem as described in \eqref{eq:regression}, where the function $f(\x\mid\w)$ is parameterized by a NN. TNs can be integrated into NNs by compressing a pretrained NN or directly training a compressed NN. A NN that incorporates TNs, either wholly or partially, is referred to as a factorized NN. The first method involves taking a readily available pretrained network \cite{tensorflow2015-whitepaper,NEURIPS2019_9015}. It then decomposes layers with a TN of choice to minimize the approximation error on the pretrained weights.
Subsequently, it is a standard practice to fine-tune the factorized network, aiming to restore any performance that may have been compromised during the decomposition process \cite{denton_exploiting_2014,lebedev_speeding-up_2015,kim_compression_2016}.
The second approach entails starting with a randomly initialized factorized network and proceeding to train it while it remains compressed. This method is recognized for improving training efficiency through the decrease in the number of parameters \cite{ye_learning_2018}.
In both methodologies, whether during fine-tuning of the first or training of the second, back-propagation is conducted based on equation \eqref{eq:regression}, with the weights represented in the form of TNs.

As an example that works for both methodologies, we show how to compress the weights for a fully connected layer with a TT decomposition following the foundational methodology of Novikov et al. \yrcite{novikov_tensorizing_2015}.
In a fully connected layer, the primary operation is the matrix-vector multiplication $\mathbf{W} \x$, involving the weight matrix $\mathbf{W} \in \mathbb{R}^{I^D \times J^D}$ and input vector $\x \in \mathbb{R}^{J^D}$. To simplify the notation, we omit the bias term. 
To prepare for applying a TN, both the weight matrix $\mathbf{W}$ and the input vector $\mathbf{x}$ must first be tensorized into higher-order tensors, resulting in $\mathbfcal{W} \in \mathbb{R}^{I \times ...  \times I \times J \times ... \times J}$ and $\mathbfcal{X} \in \mathbb{R}^{J \times ... \times J}$, respectively. The tensorization is usually performed as shown in Section \ref{sec:tensorization}.
Without loss of generality, we simplify the notation by setting $I_i=I$ and $J_j=J$ for $i,j \in {1,...,D}$.
Attaining the desired efficiency is accomplished by substituting $\mathbfcal{W}$ with a TN. Assuming a TT decomposition with uniform rank $R$, the entries of the matrix-vector product are computed as 
\begin{equation}
    \sum_{r_2 \dots r_D}
    \sum_{j_1 \dots j_D}
    w^{(1)}_{i_1 j_1 r_2} 
    w^{(2)}_{r_2 i_2 j_2 r_3}
    \cdots
    w^{(d)}_{r_D i_D j_D} 
    x_{j_1 \dots j_D} .
\end{equation}
Introducing TTs in fully connected layers reduces the computational complexity of the forward pass from $\mathcal{O}(I^D J^D)$ for the matrix representation to $\mathcal{O}(D R^2 I \max\{I^D,J^D\})$. 
The learning complexity of one backward pass is reduced from $\mathcal{O}(I^D J^D)$ to $\mathcal{O}(D^2 R^4 I \max\{I^D,J^D\})$. 
The memory complexity is reduced from $\mathcal{O}(I^D J^D)$ to $\mathcal{O}(R \max\{I^D,J^D\})$ for the forward pass and to $\mathcal{O}(R^3 \max\{I^D,J^D\})$ for the backward pass. Choosing a sufficiently small rank $R$ increases the efficiency compared to the full rank model. A theoretical guarantee for the optimal low-rank approximation, similar to the Eckhardt-Schmidt-Young theorem for matrices, does not exist for TNs \cite{vannieuwenhoven2014generic}. In Section \ref{sec:43}, we provide empirical evidence to showcase the effect of low-rank TNs on the accuracy and efficiency of NNs.

Besides TTs \cite{novikov_tensorizing_2015, garipov_ultimate_2016,tjandra_compressing_2017,wu_hybrid_2020,yang_tensor-train_2017,efthymiou_tensornetwork_2019,yu_long-term_2019,cheng_s3-net_2021}, other decompositions, e.g. Tucker \cite{kim_compression_2016,calvi_compression_2020,chu_low-rank_2021} and CP \cite{,mamalet_simplifying_2012, rigamonti_learning_2013,denton_exploiting_2014,jaderberg_speeding_2014,lebedev_speeding-up_2015,astrid_cp-decomposition_2017,chen_tensor_2020,kossaifi_factorized_2020} have been proposed for fully connected, convolutional, recurrent, and attention layers. 

\subsection{Green AI Analysis of Selected Tensor Network Models with Regard to Efficiency Metrics}\label{sec:43}
In this section, we assess how the application of TNs has led to improvements in the efficiency metrics presented in Section \ref{sec:effmetric}.
The prevalent focus in the AI community is on reporting solely accuracy rather than both accuracy and efficiency. 
This fact, combined with the absence of a standardized method for reporting efficiency, results in only a limited amount of TN-related papers that can be used in our analysis. Table \ref{sample-table} summarizes the results discussed below.  

For kernel machines, several examples the examples effectively showcase how TN can contribute towards Green AI \cite{novikov_exponential_2017,kargas_supervised_2021, wesel2021large}.
Kargas \& Sidiropoulos show that the CP decomposition model for supervised learning can match or even outperform NNs. Their research highlights comparable execution times and marginally better average accuracy for the TN methods \yrcite{kargas_supervised_2021}.
Novikov et al. report that their TN model not only offers competitive training and inference times but also yields a significant 58\% increase in accuracy compared to a NN \yrcite{novikov_exponential_2017}. Compared to the best competing method \cite{trofimov2016}, they achieved a $21 \times$ speed up with a drop in accuracy of $11.46\,\%$ \cite{novikov_exponential_2017}.
Wesel \& Batselier achieved superior accuracy $2.5$ times faster on a laptop than the best competing method \cite{hensman2013gaussian} on a cluster ($7141\,\text{s}$ vs. $18360\,\text{s}$, respectively) \yrcite{wesel2021large}. 
Compared to the uncompressed problem, using a TN reduced the number of parameters by up to $99.9\%$ \cite{wesel2021large}.\\
We highlight the contribution of TN in deep learning towards Green AI by showcasing efficiency improvements with four examples \cite{kim_compression_2016, ye_learning_2018, yin2021towards, abronin2024tqcompressor}. Kim et al. and Ye et al. have demonstrated that TNs, when applied to convolutional and recurrent layers, respectively, enhance training and inference efficiency compared to standard NNs without significantly compromising accuracy \yrcite{kim_compression_2016, ye_learning_2018}.
Across a range of datasets, Kim et al. managed significant efficiency improvements by applying TNs, achieving up to an 86.4\% reduction in memory, 79.7\% in FLOPS, 72.8\% in runtime, and 76.6\% in electricity usage while maintaining robust top-5 accuracy, reduced by no more than 2.2\% \yrcite{kim_compression_2016}.
With the application of TNs, Ye et al. managed to drastically cut down the number of parameters for the uncompressed model (by over 80,000 times), speed up convergence by 14 times, and even increase accuracy by 15.6\% \yrcite{ye_learning_2018}. 
Yin et al. proposed a framework to compress both CNNs and RNNs, which they have tested for various image classification and video recognition tasks \yrcite{yin2021towards}. For CNN models, they were able to slightly increase the accuracy compared to the uncompressed model for multiple data sets, with a compression ratio between $ 2.4 \times$ and $8.3 \times$ \cite{yin2021towards}.
Abronin et al. compressed GPT-$2_{small}$, utilizing only a fraction of the training data set to recover the performance loss. Their method outperforms competing compression approaches, achieving a $1.5\times$ compression ratio with a relative loss of $30.83\%$ in performance \yrcite{abronin2024tqcompressor}.

Employing the efficiency metrics advocated in Green AI literature clearly demonstrates the considerable potential of TNs to uphold Green AI principles, namely enhancing efficiency while maintaining accuracy.
In some scenarios, TNs enhance both accuracy and efficiency, which is an ideal outcome. However, in others, a trade-off may occur between a minor decrease in accuracy and significant improvements in energy efficiency and memory usage. 
The decision to engage in this trade-off is highly dependent on the specific context and application, making it impractical to offer a one-size-fits-all recommendation. Nevertheless, a fundamental guideline remains: a thorough analysis of accuracy and efficiency metrics is indispensable to arrive at a well-informed decision regarding the trade-off.

\begin{table*}[t]
\caption{Overview over TN models discussed in Section \ref{sec:43}. Here, accuracy is to be understood as a umbrella term \cite{schwartz2020green}. The following terms are abbreviated: classification (class.), regression (reg.), action recognition (act. reg.), Natural Language Processing (NLP), Kernel Machine (KM), Long Short-Term Memory (LSTM).}
\label{sample-table}
\vskip 0.15in
\begin{center}
\begin{small}
\begin{sc}
\begin{tabular}{lccccr}
\toprule
TN-Model & task & model & data set & \makecell{Speed up \\ Comp. ratio} & \makecell{relative change\\ in Accuracy} \\
\midrule
\cite{novikov_exponential_2017} & Class. & KM  & synthetic & $21.3 \times$ speed-up & $-\,11.46\,\%$\\
\cite{kargas_supervised_2021}  & Reg. & KM & Abalone & $1.6 \times $ speed-up & $-\,4.39\,\%$\\
\cite{wesel2021large} &  Reg.  & KM & Airline & $2.5 \times $ speed-up  &$+\,1.13\,\%$ \\
\cite{kim_compression_2016}    & Class. & AlexNet &  ImageNet & $1.8 \times$ speed-up &$+\,2.17\,\%$\\
\cite{ye_learning_2018}    & Act. Rec. & LSTM & UCF11  & $14.0 \times$ speed-up & $+\,1.13\,\%$ \\
\cite{yin2021towards} & Class.  & ResNet & CIFAR-100 & $ 2.4 \times$ comp. ratio & $+\,3.25\,\%$\\
\cite{abronin2024tqcompressor} & NLP  & GPT-2 Small & Wikitext-2 & $1.5 \times$ comp. ratio & $-\,30.83\,\%$\\
\bottomrule
\end{tabular}
\end{sc}
\end{small}
\end{center}
\vskip -0.1in
\end{table*}

\section{Critical Discussion and Conclusion}
\label{sec:conc}
This proposition paper is the first paper to propose TNs as an essential tool for realizing Green AI.
First, we addressed the exponential rise in computational demands in AI, followed by a review of the Green AI literature and various efficiency metrics suggested by researchers. A brief technical overview of TNs was provided, followed by a practical introduction to the practical implementation of TNs in deep learning and kernel machines. We demonstrated through specific examples how TNs can boost efficiency while maintaining accuracy. 

To summarize, the strength of TNs for Green AI lies in their solid mathematical foundation, logarithmic compression potential and broad applicability across different data formats and model architectures. They are a versatile tool that can significantly increase the efficiency of data preprocessing, training, and inference.\\
Regarding the three categories for Green AI, the impact of TNs can have limitations.
A limitation of TNs on AI model development, namely to enhance model efficiency, is \mbox{anticipated} for weakly correlated data since TNs are based on extending the truncated SVD to higher orders. 
In this case, other methods can be used, or TNs can be combined with other methods. So far, TNs have, for example, been effectively paired with methods such as regularization \cite{sofuoglu2020graph,wesel2021large}, automated hyperparameter search \cite{deng2022new} or knowledge distillation \cite{abronin2024tqcompressor}.

The effect of TNs on computing infrastructure and system-level impacts is clearly limited. TNs have no effect on the computing infrastructure beyond their reduced hardware requirements. Alternative low-carbon technologies are required to provide sufficient renewable energy sources, to reduce the energy demand of data centres and the emissions associated with manufacturing, transporting and recycling hardware. 
Potential system-level impacts of TNs (and of most methods that increase the efficiency of AI models) include the acceleration of carbon-intensive technologies and lifestyle choices or the rebound effect. This effect occurs when the expected gains are diminished due to increased technology usage. 
However, the rebound effect is not exclusively detrimental. It can enhance inclusivity by lowering barriers to entry for previously excluded researchers and, therefore, increase usage. 
This broader inclusion has the potential to bring together a more diverse group of researchers, which could enrich the progress of AI.
Lastly, the role of AI in addressing climate change issues, as discussed in works like Huntingford et al., must be considered \yrcite{huntingford2019machine}. 
Mitigating undesired system-level impacts requires, e.g., policy considerations and technology management.
Ensuring that AI solutions foster sustainability requires a careful appraisal of the costs and benefits, a consideration that, while important, is beyond the scope of this paper. 

Regarding the three pillars of sustainability, TNs offer notable potential benefits. First, by reducing the need for hardware and computational resources, TNs can cut costs, making them an economically more sustainable solution and a viable choice for industrial applications.
Second, the reduced hardware and computational demands of TNs decrease reliance on expensive, potentially external computing resources. As discussed above, reduced hardware requirements can lower entry barriers to AI research, potentially fostering inclusivity and social sustainability. Additionally, the capacity to process sensitive data on-site can enhance data privacy.
Third, the ability of TNs to minimize hardware use and shorten computational time can significantly decrease the embodied and operational emissions associated with AI. Consequently, the widespread adoption of TNs in AI models could profoundly affect the environmental sustainability of AI.
Looking forward, as we approach the constraints imposed by Moore's Law, the quest for algorithmic innovation is anticipated to increasingly focus on enhancing efficiency. 
In this scenario, TNs, renowned for their logarithmic compression potential, can play a crucial role in facilitating continued advancement of AI.

A potential direction for future work could be to provide a comprehensive overview and assessment of particularly beneficial combinations of additional Green AI methods with TNs, considering application-specific requirements. Regarding TNs specifically, there are open challenges that can dampen their potential benefits on Green AI. For example, implementing favourable design choices for TNs often requires experience. Practitioners' guidelines or theoretical frameworks to support decision-making are yet to be developed. A first step in that direction could be quantifying TNs’ effect on the whole model development phase and the savings in embodied emissions.\\
We advocate for optimizing TNs to Green AI objectives. That includes intensifying research on data- or hardware-efficient TN models, training TN models from scratch, or gradually compressing TN models while training. Open questions evolve around tensor rank approximation. Finding the true tensor rank is an NP-hard problem. Furthermore, for tensors, there exists no theorem akin to the Schmidt-Eckart-Young theorem to provide a theoretical guarantee for the optimal low-rank approximation. Moreover, future research could further investigate proper weight initialization or efficient hyperparameter training, among others.

We envision that TNs will have a broad application in diverse sectors of AI and contribute to more sustainable AI. We encourage researchers to integrate TNs in their research, address the open questions, and advance AI progress. 

\section*{Acknowledgements}
We would like to thank the anonymous reviewers for their numerous suggestions and improvements which have greatly improved the quality
of this paper. Eva Memmel and Frederiek Wesel, and thereby this work, are supported by the Delft University of Technology AI Labs program. The authors declare no competing
interests.

\section*{Impact Statement}
This paper links tensor networks and Green AI research fields to enhance AI model sustainability and research inclusivity. The potential societal impact is discussed throughout the paper, especially in Section \ref{sec:conc}.

\bibliography{example_paper}
\bibliographystyle{icml2024}
\end{document}